\documentclass[fleqn,10pt]{wlscirep}
\usepackage[T1]{fontenc}
\usepackage{lmodern}
\usepackage{amsmath,amssymb}
\usepackage{booktabs}
\usepackage{multirow}
\usepackage{subfig}
\usepackage{hyperref}

\title{KANMixer: a minimal KAN-centered mixer for long-term time series forecasting}

\author[1,+]{Lingyu Jiang}
\author[1]{Dengzhe Hou}
\author[2]{Yuping Wang}
\author[5]{Yao Su}
\author[3]{Shuo Xing}
\author[3]{Wenjing Chen}
\author[4]{Xin Zhang}
\author[3]{Zhengzhong Tu}
\author[5]{Ziming Zhang}
\author[1,3,5,+]{Fangzhou Lin}
\author[1]{Michael Zielewski}
\author[1]{Kazunori Yamada}

\affil[1]{Graduate School of Information Sciences, Tohoku University, Sendai 980-8579, Japan}
\affil[2]{University of Michigan, Ann Arbor, MI 48109, USA}
\affil[3]{Texas A\&M University, College Station, TX 77843, USA}
\affil[4]{San Diego State University, San Diego, CA 92182, USA}
\affil[5]{Worcester Polytechnic Institute, Worcester, MA 01609, USA}
\affil[+]{Corresponding author:
\href{mailto:jiang.lingyu.p7@dc.tohoku.ac.jp}{jiang.lingyu.p7@dc.tohoku.ac.jp},
\href{mailto:flin2@wpi.edu}{flin2@wpi.edu}}
\keywords{long-term time series forecasting; Kolmogorov-Arnold network; spline basis function; temporal mixing; deep learning}

\begin{abstract}
Long-term time series forecasting (LTSF) underpins critical applications from energy management to weather prediction, yet achieving reliable multi-step-ahead accuracy remains challenging. Existing LTSF approaches, dominated by MLP- and Transformer-based architectures, either rely on simple linear mappings or introduce increasingly complex hand-crafted inductive biases, raising the question of whether a more expressive and principled nonlinear core could offer a better alternative. Therefore, we investigate whether Kolmogorov-Arnold Networks (KANs),  a recently proposed model featuring adaptive basis functions capable of granular modulation of nonlinearities, can improve LTSF performance, and under which design choices they are most effective. Specifically, we propose KANMixer, a minimal KAN-centered architecture consisting of a multi-scale pooling frontend, a KAN-based temporal mixing backbone, and prediction heads. By avoiding heavy auxiliary modules, KANMixer enables a clear assessment of KAN components in LTSF. Across 28 benchmark–horizon settings against nine baselines, KANMixer achieves the best MSE in 16 settings and the best MAE in 11. Furthermore, extensive ablations on three representative datasets show that KAN effectiveness depends strongly on the choice of edge function; B-spline bases outperform Fourier and Wavelet alternatives; the prediction head contributes most to the gains; moderate depth is preferred over deeper unstable stacks; and decomposition priors help MLP but harm KAN. Beyond practical guidance for integrating KAN into LTSF, these results reveal an underexplored dependency between structural priors and backbone nonlinearity: design choices that benefit MLP can degrade KAN.
\end{abstract}

\begin{document}

\flushbottom
\maketitle
\thispagestyle{empty}

% ============================================================
\section*{Introduction}
% ============================================================

Long-term time series forecasting (LTSF) is a fundamental task with wide scientific and societal applications. Reliable multi-step-ahead predictions of electricity demand support grid stability and renewable energy integration~\cite{en13246623}. Similarly, weather forecasting accuracy~\cite{weather} directly affects agricultural planning and disaster preparedness, while traffic flow prediction~\cite{traffic} underpins intelligent transportation systems. In all these domains, forecasting errors over horizons of days to weeks carry real costs, motivating continued improvement in model accuracy and reliability~\cite{ltsf}. Therefore, understanding which architectural components drive forecasting performance, and under what conditions, is a question of both scientific interest and practical significance.

Deep learning has largely supplanted traditional statistical methods such as ARIMA~\cite{arima} and classical machine learning~\cite{xgboost,ke2017lightgbm}, with model architectures evolving from recurrent neural networks~\cite{lstm,d2021developing} and graph neural networks~\cite{gnn1,wu2021inductive,jiang2025time} to Transformer-based models~\cite{vaswani2017attention} such as Informer~\cite{informer}, Autoformer~\cite{autoformer}, FEDformer~\cite{fedformer}, PatchTST~\cite{patchtst}, iTransformer~\cite{liu2024itransformer}, and Crossformer~\cite{zhang2023crossformer}. However, DLinear~\cite{dlinear} showed that a simple linear model can outperform complex Transformers on many benchmarks, motivating the search for more principled modeling cores. Subsequent MLP-based models, including TSMixer~\cite{tsmixer}, TiDE~\cite{tide}, FreTS~\cite{yi2023frequencydomain}, TimesNet~\cite{wu2023timesnet}, TimeMixer~\cite{wang2024timemixer}, and SOFTS~\cite{softs}, introduced hand-crafted structural priors such as frequency decomposition~\cite{fredf} and patch mixing~\cite{patchmixer}. However, recent studies suggest that the gains from these increasingly complex designs are often limited or inconsistent under controlled evaluation settings~\cite{position}.

These observations suggest that the backbone nonlinearity itself may be a primary bottleneck. Kolmogorov-Arnold Networks (KAN)~\cite{liu2025kan}, inspired by the Kolmogorov-Arnold representation theorem~\cite{kart}, offer an alternative worth examining: KAN replaces fixed-form nonlinearities with trainable B-spline basis functions~\cite{kainn}, enabling fine-grained local modulation and compact universal approximation. Recent KAN variants have addressed efficiency trade-offs through alternative bases~\cite{chebykan,kaf,wavkan}, and a theoretical analysis shows that B-spline KAN is closely related to radial basis function networks~\cite{kan2rbf}. These developments motivate a systematic investigation of KAN in LTSF, particularly regarding when KAN components are beneficial and where their gains primarily arise.

\begin{figure}[t]
\centering
\includegraphics[width=0.48\linewidth]{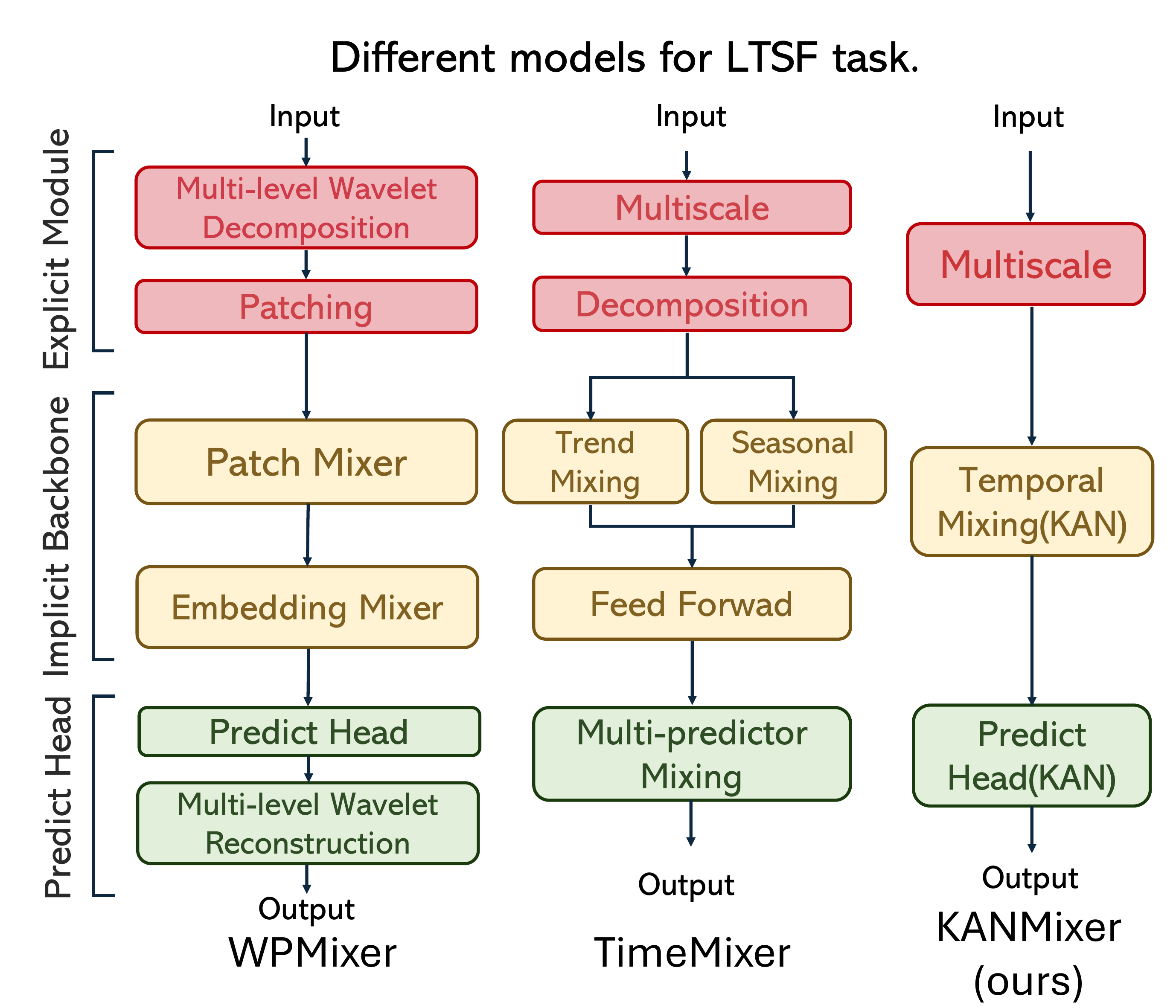}
\caption{Architectural comparison of representative LTSF models. Compared with prior designs that combine explicit modules such as decomposition or patching with implicit mixing blocks, KANMixer adopts a more concise architecture with a multi-scale frontend, a KAN-based temporal mixing backbone, and KAN-based prediction heads.}
\label{fig1}
\end{figure}

Prior work has begun integrating KAN into LTSF. TimeKAN~\cite{huang2025timekan} embeds multi-order KAN within a frequency decomposition pipeline and achieves competitive results. RMoK~\cite{han2025kanseffectivemultivariatetime} validates KAN as a building block in a mixture-of-experts framework. However, when KAN is embedded inside a decomposition-heavy pipeline, our ablations suggest it becomes nearly interchangeable with MLP, as the surrounding architecture may account for much of the observed performance. This observation calls for a minimal design that isolates KAN's contribution. We introduce KANMixer, a minimal KAN-centered architecture with three modules: a multi-scale pooling frontend, a KAN-based temporal mixing backbone, and KAN-based prediction heads. The design deliberately avoids decomposition and other heavy engineering modules, enabling controlled attribution of performance gains to KAN components (Fig.~\ref{fig1}).

In a mixed comparison against nine baselines spanning KAN-based, Transformer-based, MLP-based, CNN-based, and foundation model architectures, KANMixer records the lowest numerical MSE in 16 of 28 configurations. Beyond benchmarking, we conduct five ablation studies on three representative datasets to diagnose where KAN contributes most, how sensitive performance is to basis-function choice, and how decomposition priors interact with KAN. A counterintuitive finding is that decomposition, which consistently helps MLP, degrades KAN performance in our experiments. These analyses inform concrete design recommendations for KAN-based LTSF models. The main contributions of this paper are as follows:

\begin{itemize}
\item We propose KANMixer, a structurally simple model featuring KAN as its modeling core. KANMixer achieves competitive performance compared with more complex SOTA models, demonstrating its effectiveness.

\item We provide a systematic analysis of KAN's modeling characteristics in LTSF, showing that its performance gains are closely related to the adaptive flexibility of its basis functions. Our analysis also indicates that structural priors interact differently with KAN than with MLP.

\item To our knowledge, we provide one of the first empirically grounded sets of practical guidelines for effectively applying KAN to LTSF, emphasizing the critical importance of the prediction head and optimal network depth in maximizing forecasting performance.
\end{itemize} 

% ============================================================
\section*{Results}
% ============================================================

\subsection*{Experimental setup}

\noindent\textbf{Datasets.} We evaluate on seven standard LTSF benchmarks: ETTh1, ETTh2, ETTm1, ETTm2 (ETT series~\cite{informer}), Exchange Rate, Weather, and Electricity. The ETT datasets use a 6:2:2 train/validation/test split; the remaining datasets use 7:1:2.

\noindent\textbf{Baselines.} We compare against nine methods: TimeKAN~\cite{huang2025timekan}, iTransformer~\cite{liu2024itransformer}, PatchTST~\cite{patchtst}, TimeMixer~\cite{wang2024timemixer}, DLinear~\cite{dlinear}, FreTS~\cite{yi2023frequencydomain}, TiDE~\cite{tide}, TimesNet~\cite{wu2023timesnet}, and Time-FFM~\cite{liu2024timeffm}.

\noindent\textbf{Evaluation.} We report MSE and MAE~\cite{autoformer,mse} over four prediction horizons $P \in \{96, 192, 336, 720\}$ with look-back window $L = 96$. KANMixer, TimeKAN, and TimeMixer results are averaged over five independent runs; remaining baselines use paper-reported values.

\subsection*{Main forecasting results}

Table~\ref{table1} reports forecasting performance across all datasets. We distinguish two levels of comparison throughout this section. Controlled comparisons involve KANMixer, TimeKAN, and TimeMixer, all reproduced in our environment with identical splits, RevIN normalization, and evaluation code, and averaged over five independent runs. Within this controlled set, KANMixer achieves lower MSE than both TimeKAN and TimeMixer in 14 of 28 settings, with mean differences of 0.010--0.020 MSE on ETTh1 and 0.005--0.015 on ETTm1/ETTm2; these differences consistently exceed the within-model standard deviations ($\leq$0.005 MSE) and therefore likely reflect a reliable performance advantage. Contextual comparisons involve the remaining seven baselines (iTransformer, PatchTST, Time-FFM, TimesNet, FreTS, DLinear, TiDE) whose results are taken from their respective publications; these may be affected by differences in hardware, software environment, seed selection, and normalization conventions. Among all 28 settings, KANMixer records the lowest numerical MSE in 16 and lowest MAE in 11 configurations.

The advantage is most pronounced on ETTh1, where KANMixer improves MSE by an average of 4.9\% across horizons. On ETTm1 and ETTm2, KANMixer leads in 3 of 4 horizon configurations each. On Electricity (321 variables), iTransformer leads due to its explicit cross-variate attention; on Exchange Rate, Time-FFM benefits from cross-dataset pre-training. These boundary conditions are consistent with KANMixer's channel-independent design.

\begin{table*}[t]
  \centering
  \footnotesize
  \setlength{\tabcolsep}{0.9mm}
  \renewcommand\arraystretch{1.02}
  \resizebox{\textwidth}{!}{%
  \begin{tabular}{ll*{10}{cc}}
    \toprule
    \multirow{3}{*}{\textbf{Models}} & \multirow{3}{*}{$ $} &
      \multicolumn{2}{c}{\textbf{KANMixer}} &
      \multicolumn{2}{c}{TimeKAN} &
      \multicolumn{2}{c}{TimeMixer} &
      \multicolumn{2}{c}{iTransformer} &
      \multicolumn{2}{c}{Time-FFM} &
      \multicolumn{2}{c}{TimesNet} &
      \multicolumn{2}{c}{PatchTST} &
      \multicolumn{2}{c}{FreTS} &
      \multicolumn{2}{c}{DLinear} &
      \multicolumn{2}{c}{TiDE} \\[.2ex]
      & &
      \multicolumn{2}{c}{\textbf{(Ours)}} &
      \multicolumn{2}{c}{2025} &
      \multicolumn{2}{c}{2024} &
      \multicolumn{2}{c}{2024} &
      \multicolumn{2}{c}{2024} &
      \multicolumn{2}{c}{2023} &
      \multicolumn{2}{c}{2023} &
      \multicolumn{2}{c}{2024} &
      \multicolumn{2}{c}{2023} &
      \multicolumn{2}{c}{2023} \\[-.2ex]
      \cmidrule(lr){3-4}\cmidrule(lr){5-6}\cmidrule(lr){7-8}
      \cmidrule(lr){9-10}\cmidrule(lr){11-12}\cmidrule(lr){13-14}
      \cmidrule(lr){15-16}\cmidrule(lr){17-18}\cmidrule(lr){19-20}
      \cmidrule(lr){21-22}
      & & MSE & MAE & MSE & MAE & MSE & MAE & MSE & MAE
        & MSE & MAE & MSE & MAE & MSE & MAE & MSE & MAE
        & MSE & MAE & MSE & MAE \\
    \midrule
   \multirow{4}{*}{ETTh1}
       & 96 & \textbf{0.367} & \textbf{0.392} & \underline{0.384} & \underline{0.396} & 0.385 & 0.402 & 0.386 & 0.405 & 0.385 & 0.400 & \underline{0.384} & 0.402 & 0.460 & 0.447 & 0.395 & 0.407 & 0.397 & 0.412 & 0.479 & 0.464 \\
     & 192 & \textbf{0.422} & \underline{0.427} & 0.437 & \textbf{0.425} & \underline{0.436} & 0.429 & 0.441 & 0.436 & 0.439 & 0.430 & 0.439 & 0.429 & 0.512 & 0.477 & 0.490 & 0.477 & 0.446 & 0.441 & 0.525 & 0.492 \\
      & 336 & \textbf{0.446} & \underline{0.444} & \underline{0.476} & \textbf{0.439} & 0.529 & 0.456 & 0.487 & 0.458 & 0.480 & 0.449 & 0.638 & 0.469 & 0.546 & 0.496 & 0.510 & 0.480 & 0.489 & 0.467 & 0.565 & 0.515 \\
      & 720 & \textbf{0.442} & \textbf{0.455} & 0.468 & 0.470 & 0.483 & 0.474 & 0.503 & 0.491 & \underline{0.462} & \underline{0.456} & 0.512 & 0.500 & 0.544 & 0.517 & 0.568 & 0.538 & 0.513 & 0.510 & 0.594 & 0.558 \\
      \midrule
    \multirow{4}{*}{ETTh2}
      &   96 & \textbf{0.288} & \underline{0.342} & 0.306 & 0.353 & \underline{0.289} & \textbf{0.341} & 0.297 & 0.349 & 0.301 & 0.351 & 0.340 & 0.374 & 0.308 & 0.355 & 0.332 & 0.364 & 0.340 & 0.394 & 0.400 & 0.440 \\
      & 192 & \textbf{0.371} & \underline{0.394} & \underline{0.375} & \textbf{0.392} & 0.391 & 0.403 & 0.380 & 0.400 & 0.378 & 0.397 & 0.402 & 0.414 & 0.393 & 0.405 & 0.451 & 0.457 & 0.482 & 0.479 & 0.528 & 0.509 \\
      & 336 & \textbf{0.419} & 0.433 & 0.425 & 0.435 & 0.426 & 0.433 & 0.428 & \underline{0.432} & \underline{0.422} & \textbf{0.431} & 0.452 & 0.452 & 0.427 & 0.436 & 0.466 & 0.473 & 0.591 & 0.541 & 0.643 & 0.571 \\
      & 720 & 0.448 & 0.454 & 0.471 & 0.464 & 0.468 & 0.468 & \textbf{0.427} & \underline{0.445} & \textbf{0.427} & \textbf{0.444} & 0.462 & 0.468 & \underline{0.436} & 0.450 & 0.485 & 0.471 & 0.839 & 0.661 & 0.874 & 0.679 \\
      \midrule
    \multirow{4}{*}{ETTm1}
       & 96 & \textbf{0.311} & \textbf{0.355} & 0.326 &\underline{0.363} & \underline{0.320} & 0.360 & 0.334 & 0.368 & 0.336 & 0.369 & 0.338 & 0.375 & 0.352 & 0.374 & 0.337 & 0.374 & 0.346 & 0.374 & 0.364 & 0.387 \\
      & 192 & \textbf{0.357} & \textbf{0.378} & \underline{0.359} & \underline{0.384} & 0.370 & 0.387 & 0.377 & 0.391 & 0.378 & 0.389 & 0.374 & 0.387 & 0.390 & 0.393 & 0.382 & 0.398 & 0.382 & 0.391 & 0.398 & 0.404 \\
      & 336 & \textbf{0.381} & \textbf{0.400} & 0.390 & 0.407 & \underline{0.389} & \underline{0.402} & 0.426 & 0.420 & 0.411 & 0.410 & 0.410 & 0.411 & 0.421 & 0.410 & 0.420 & 0.423 & 0.415 & 0.451 & 0.428 & 0.425 \\
      & 720 & \underline{0.444} & \underline{0.439} & \textbf{0.442} & \textbf{0.435} & 0.451 & \underline{0.439} & 0.491 & 0.459 & 0.469 & 0.441 & 0.478 & 0.450 & 0.462 & 0.449 & 0.490 & 0.471 & 0.473 & 0.451 & 0.487 & 0.461 \\
      \midrule
    \multirow{4}{*}{ETTm2} & 96 & \textbf{0.173} & \textbf{0.257} & 0.177 & \underline{0.259} & \underline{0.176} & \textbf{0.257} & 0.180 & 0.264 & 0.181 & 0.267 & 0.187 & 0.267 & 0.183 & 0.270 & 0.186 & 0.275 & 0.193 & 0.293 & 0.207 & 0.305 \\
      & 192 & \textbf{0.239} & \underline{0.303} & 0.242 & 0.304 & \underline{0.240} & \textbf{0.302} & 0.250 & 0.309 & 0.247 & 0.308 & 0.249 & 0.309 & 0.255 & 0.315 & 0.259 & 0.323 & 0.284 & 0.361 & 0.290 & 0.364 \\
      & 336 & \textbf{0.300} & \textbf{0.343} & 0.304 & \underline{0.344} & \underline{0.303} & \textbf{0.343} & 0.311 & 0.348 & 0.309 & 0.347 & 0.321 & 0.351 & 0.309 & 0.347 & 0.420 & 0.423 & 0.382 & 0.429 & 0.377 & 0.422 \\
      & 720 & \textbf{0.398} & \textbf{0.401} & \underline{0.400} & \textbf{0.401} & 0.404 & 0.404 & 0.412 & 0.407 & 0.406 & 0.404 & 0.408 &  \underline{0.403} & 0.412 & 0.404 & 0.559 & 0.511 & 0.558 & 0.525 & 0.558 & 0.524 \\
      \midrule
   \multirow{4}{*}{Exchange} & 96 & \underline{0.083} & \underline{0.202} & 0.086 & 0.206 & 0.090 & 0.235 & 0.086 & 0.206 & \textbf{0.081} & \textbf{0.201} & 0.107 & 0.234 & 0.088 & 0.205 & 0.093 & 0.220 & 0.088 & 0.218 & 0.094 & 0.218 \\
      & 192 & \underline{0.174} & \underline{0.297} & 0.182 & 0.303 & 0.187 & 0.343 & 0.177 & 0.299 & \textbf{0.168} & \textbf{0.293} & 0.226 & 0.344 & 0.176 & 0.299 & 0.222 & 0.350 & 0.176 & 0.315 & 0.184 & 0.307 \\
      & 336 & 0.323 & 0.411 & 0.349 & 0.427 & 0.353 & 0.473 & 0.331 & 0.417 & \textbf{0.299} & \textbf{0.396} & 0.367 & 0.448 & \underline{0.301} & \underline{0.397} & 0.386 & 0.467 & 0.313 & 0.427 & 0.349 & 0.431 \\
      & 720 & 0.841 & \underline{0.687} & 0.923 & 0.719 & 0.934 & 0.761 & 0.847 & 0.691 & \textbf{0.805} & \textbf{0.674} & 0.964 & 0.746 & 0.901 & 0.714 & 0.875 & 0.708 & \underline{0.839} & 0.695 & 0.852 & 0.698 \\
      \midrule
     \multirow{4}{*}{Weather} & 96 & \textbf{0.162} & \textbf{0.209} & \underline{0.163} & \textbf{0.209} & \textbf{0.162} & \textbf{0.209} & 0.174 & \underline{0.214} & 0.191 & 0.230 & 0.172 & 0.220 & 0.186 & 0.227 & 0.171 & 0.227 & 0.195 & 0.252 & 0.202 & 0.261 \\
      & 192 & \textbf{0.206} & \textbf{0.249} & \underline{0.209} & \underline{0.252} & 0.211 & 0.254 & 0.221 & 0.254 & 0.236 & 0.267 & 0.219 & 0.261 & 0.234 & 0.265 & 0.218 & 0.280 & 0.237 & 0.295 & 0.242 & 0.298 \\
      & 336 & 0.264 & \textbf{0.291} & 0.264 & \underline{0.292} & \underline{0.263} & 0.293 & 0.278 & 0.296 & 0.289 & 0.303 & \textbf{0.246} & 0.337 & 0.284 & 0.301 & 0.265 & 0.317 & 0.282 & 0.331 & 0.287 & 0.335 \\
      & 720 & 0.345 & \underline{0.344} & \underline{0.340} & \textbf{0.343} & 0.344 & 0.348 & 0.358 & 0.347 & 0.362 & 0.350 & 0.365 & 0.359 & 0.356 & 0.349 & \textbf{0.326} & 0.351 & 0.345 & 0.382 & 0.351 & 0.386 \\
      \midrule
   \multirow{4}{*}{Electricity} & 96 & 0.162 & 0.260 & 0.177 & 0.267 & \underline{0.156} & \underline{0.247} & \textbf{0.148} & \textbf{0.240} & 0.198 & 0.282 & 0.168 & 0.272 & 0.190 & 0.296 & 0.171 & 0.260 & 0.210 & 0.302 & 0.237 & 0.329 \\
      & 192 & 0.171 & 0.261 & 0.182 & 0.272 & \underline{0.166} & \underline{0.257} & \textbf{0.162} & \textbf{0.253} & 0.199 & 0.285 & 0.184 & 0.322 & 0.199 & 0.304 & 0.177 & 0.268 & 0.210 & 0.305 & 0.236 & 0.330 \\
      & 336 & 0.191  & 0.283 & 0.198 & 0.287 & \underline{0.185} & \underline{0.275} & \textbf{0.178} & \textbf{0.269} & 0.212 & 0.298 & 0.198 & 0.300 & 0.217 & 0.319 & 0.190 & 0.284 & 0.223 & 0.319 & 0.249 & 0.344 \\
      & 720 & 0.229 & \underline{0.313} & 0.239 & 0.321 & \underline{0.224} & \textbf{0.312} & 0.225 & 0.317 & 0.253 & 0.330 & \textbf{0.220} & 0.320 & 0.258 & 0.352 & 0.228 & 0.316 & 0.258 & 0.350 & 0.284 & 0.373 \\
      \midrule
    \multicolumn{2}{l}{1st Count} &
       \textbf{16}& \textbf{11} & 1 & 7 & 1 & 6 & 4 & 3 & 5
        & 6 & 2 & 0 & 0 & 0 & 1 & 0 & 0 & 0 & 0& 0\\
    \bottomrule
  \end{tabular}%
  }
  \caption{Forecasting results with look-back window $L=96$ and prediction lengths $P \in \{96, 192, 336, 720\}$. Best result in \textbf{bold}; second-best \underline{underlined}.}
  \label{table1}
\end{table*}

\subsection*{Comparison with strong recent baselines}

To better understand where KANMixer's advantages and limitations lie, we compare it in detail with the two strongest individual baselines: iTransformer and Time-FFM. iTransformer leads on the Electricity dataset (321 variables) through explicit cross-variate attention, a mechanism structurally absent in KANMixer's channel-independent design. On the six remaining datasets (ETTh1, ETTh2, ETTm1, ETTm2, Weather, Exchange), KANMixer outperforms iTransformer in the majority of configurations, with an average 4.9\% MSE improvement on ETTh1. Time-FFM, a large pre-trained foundation model, excels on Exchange Rate where cross-dataset pre-training provides macroeconomic priors unavailable to single-dataset models. KANMixer outperforms Time-FFM on 5 of 7 datasets using only 321.73K parameters. The first-place counts are spread across multiple datasets rather than concentrated on one or two. This suggests broad competitiveness rather than a narrow dataset-specific advantage: 16 MSE / 11 MAE versus Time-FFM's 5/6 and iTransformer's 4/3.

\subsection*{Ablation Studies}

\begin{figure*}[t]
\centering
\includegraphics[width=0.98\linewidth]{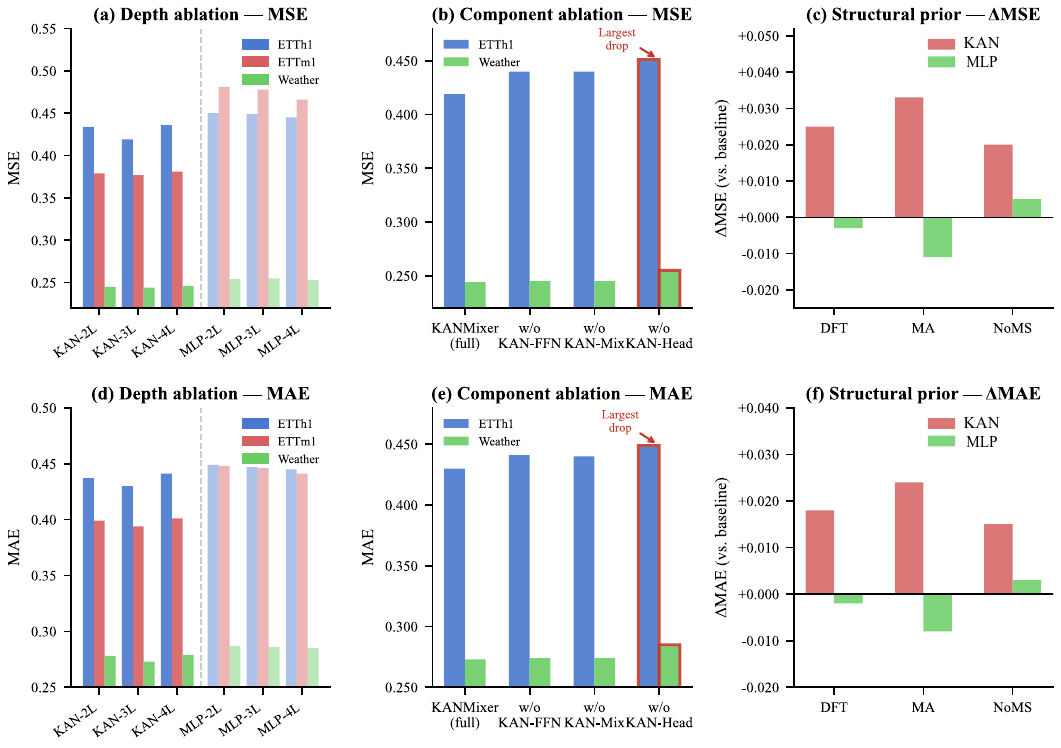}
\caption{Ablation analyses of KANMixer. (a,d) Depth ablation of KAN and MLP backbones with 2--4 layers on ETTh1, ETTm1, and Weather. The 3-layer KAN variant achieves the best performance, while MLP remains consistently inferior. (b,e) Component-wise ablation by replacing each KAN module with its MLP counterpart; the prediction head is the most critical component, as its removal leads to the largest degradation. (c,f) Ablation of structural priors and multi-scale design, measured as performance change relative to the baseline. Adding DFT or moving-average decomposition, or removing the multi-scale module, consistently harms KAN and has smaller or mixed effects on the MLP counterpart.}
\label{fig2}
\end{figure*}

\begin{figure*}[t!]
  \centering
  \small
  \subfloat[ETTh1\label{fig:ETTh1}]{
    \includegraphics[width=0.48\linewidth]{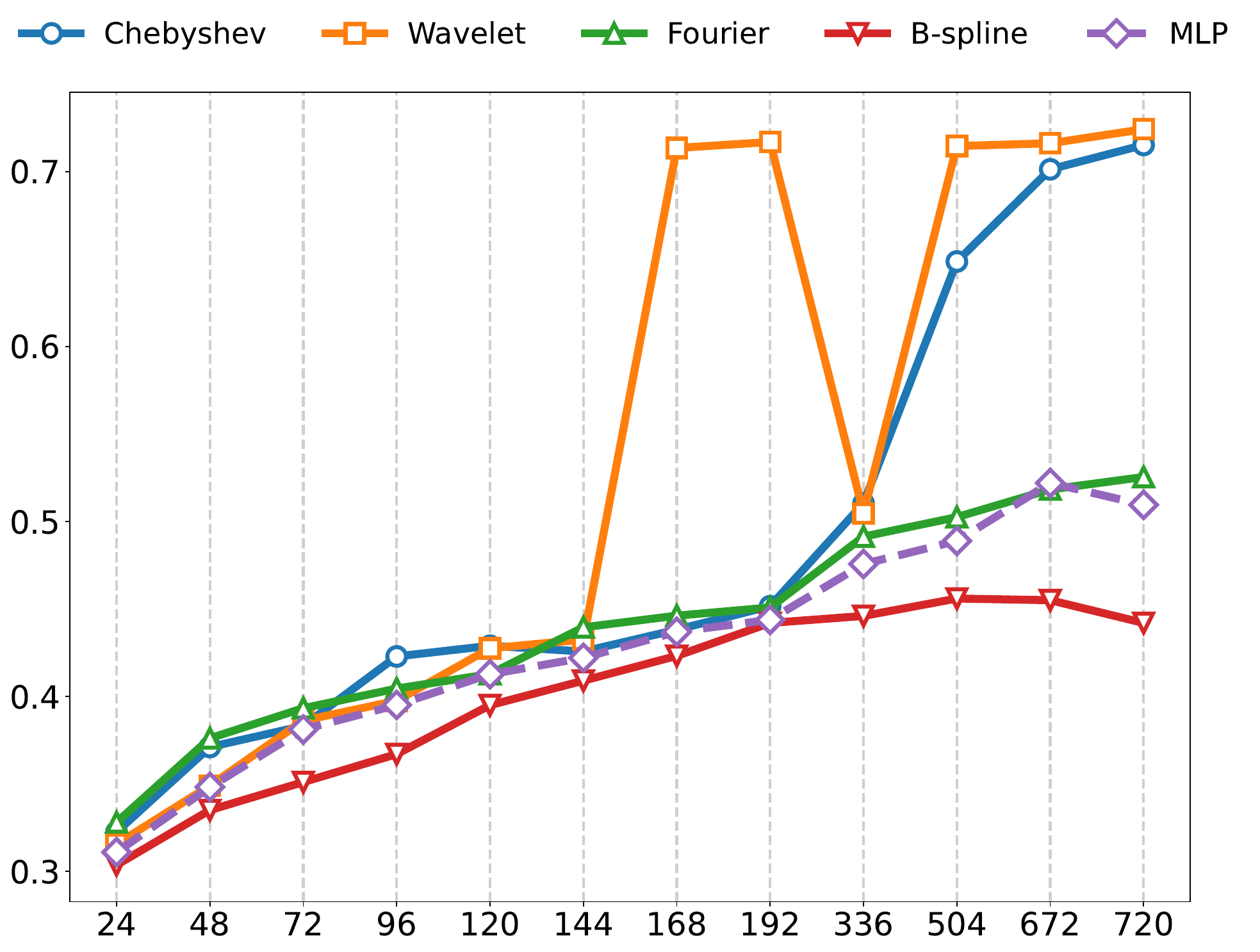}
  }
  \vspace{4pt}
  \subfloat[ETTm1\label{fig:ETTm1}]{
    \includegraphics[width=0.48\linewidth]{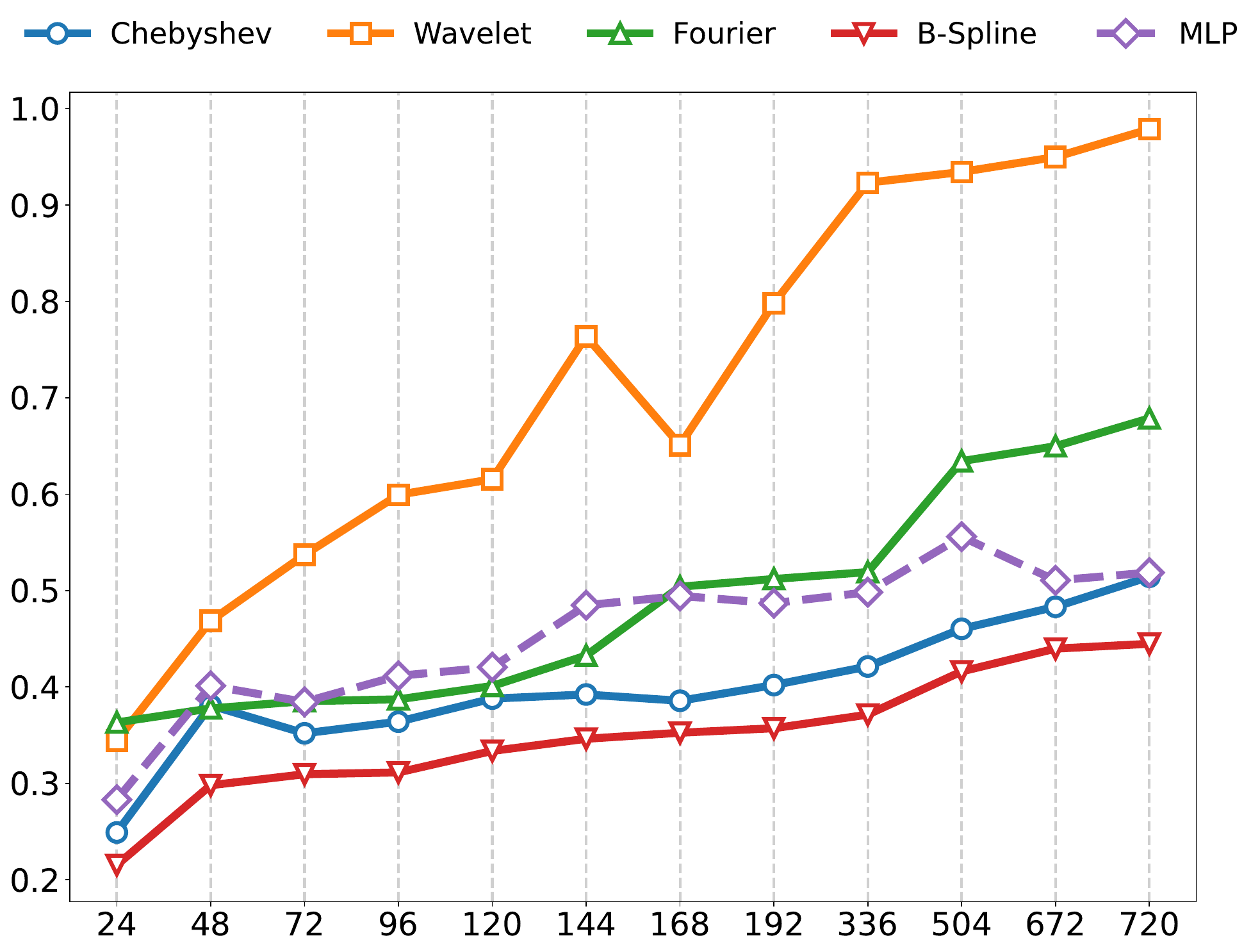}
  }
  \caption{MSE vs. prediction length for KANMixer variants using different basis functions on ETTh1 and ETTm1. B-spline is the only variant that consistently improves over MLP.}
  \label{fig:kan_variants}
\end{figure*}

We conduct ablation studies to answer four questions: (1) whether the performance gain comes from using KAN rather than an MLP backbone, and how sensitive this gain is to depth; (2) which KAN component contributes most within KANMixer; (3) how the choice of basis function affects forecasting performance; and (4) how KAN interacts with explicit structural priors and the multi-scale design. The results are summarized in Fig.~\ref{fig2} and Fig.~\ref{fig:kan_variants}.

\noindent\textbf{Depth ablation: KAN vs. MLP}

A central question is whether the advantage of KAN observed in other domains also transfers to LTSF~\cite{tran2024}. Prior ablations in TimeKAN reported only minor differences between KAN and MLP, which is consistent with our hypothesis that embedding a Chebyshev-basis KAN into a decomposition-heavy pipeline may weaken the adaptive advantage of KAN. To examine this directly, we replace the KAN layers in KANMixer with MLP layers and compare 2-, 3-, and 4-layer variants, with the width of each model independently tuned. As shown in Fig.~\ref{fig2}(a,d), KAN consistently outperforms MLP across all three datasets, and the 3-layer KAN variant gives the best overall performance. Deeper KAN stacks, however, exhibit mild instability, suggesting that increasing depth does not further improve forecasting accuracy in this setting.

%\begin{table}[t]
%\centering
%\footnotesize
%\begin{tabular}{lcccccc}
%\hline
%\multirow{2}{*}{Model} & \multicolumn{2}{c}{ETTh1}
%                       & \multicolumn{2}{c}{ETTm1}
%                       & \multicolumn{2}{c}{Weather} \\
%                       & MSE & MAE & MSE & MAE & MSE & MAE \\ \hline
%KAN-2L    & 0.434 & 0.437 & 0.379 & 0.396 & 0.245 & 0.278 \\
%KAN-3L    & \textbf{0.419} & \textbf{0.430} & \textbf{0.377} & \textbf{0.394} & %\textbf{0.244} & \textbf{0.273} \\
%KAN-4L    & 0.436 & 0.438 & 0.381 & 0.396 & 0.246 & 0.289 \\
%MLP-2L & 0.450 & 0.458 & 0.481 & 0.516 & 0.254 & 0.284 \\
%MLP-3L & 0.449 & 0.445 & 0.478 & 0.514 & 0.255 & 0.285 \\
%MLP-4L & 0.445 & 0.445 & 0.466 & 0.504 & 0.253 & 0.284 \\ \hline
%\end{tabular}
%\caption{Depth ablation comparing KAN vs. MLP within KANMixer. Width independently tuned per variant via grid search over $d_{\text{model}} \in \{32,64,128\}$. Training restarts (divergence criterion) occurred in ${<}5\%$ of KAN-4L runs and were not observed for KAN-3L or MLP variants.}
%\label{tab:ablation_layers}
%\end{table}

\noindent\textbf{Component-wise ablation}
\label{sec:component}

To identify which KAN component contributes most to the overall performance, we replace each KAN module with its MLP counterpart while keeping the rest of the architecture unchanged. As shown in Fig.~\ref{fig2}(b,e), the KAN prediction head is the most critical component: replacing it leads to the largest degradation among all module-wise substitutions. This suggests that the final mapping from latent representations to multi-step forecasts benefits most from the adaptive nonlinearity provided by KAN, whereas the gains from KAN-based mixing and feed-forward transformation are comparatively smaller. From a practical perspective, this result indicates that replacing only the prediction head with a KAN layer may already be a simple yet effective modification for existing forecasting architectures.

%\begin{table}[t]
%\centering
%\footnotesize
%\begin{tabular}{lcccc}
%\hline
%\multirow{2}{*}{Model} & \multicolumn{2}{c}{ETTh1}
%                       & \multicolumn{2}{c}{Weather} \\
%                       & MSE & MAE & MSE & MAE \\ \hline
%KANMixer (ours)       & \textbf{0.419} & \textbf{0.430} & \textbf{0.244} & \textbf{0.273} \\
%w/o KAN-FFN           & 0.440 & 0.441 & 0.245 & 0.274 \\
%w/o KAN-Mixing        & 0.440 & 0.435 & 0.245 & 0.274 \\
%w/o KAN-Prediction    & 0.451 & 0.439 & 0.255 & 0.278 \\ \hline
%\end{tabular}
%\caption{Component ablation: each KAN module replaced individually by its MLP counterpart.}
%\label{tab:ablation_components}
%\end{table}

\noindent\textbf{Basis function ablation}
\label{sec:basis}

We compare four basis function variants together with an MLP baseline (Fig.~\ref{fig:kan_variants}). B-spline is the only basis that consistently outperforms MLP across all prediction lengths. Chebyshev performs less consistently and degrades at longer horizons on ETTh1. Fourier and Wavelet both underperform MLP across most settings, with Wavelet showing pronounced instability at longer horizons. These results indicate that basis function choice is a primary design variable for KAN in LTSF, and suggest that the local support of B-spline may be an important factor behind its advantage. These findings also offer a possible explanation for the observation in TimeKAN that KAN and MLP perform similarly. When a Chebyshev basis is embedded within a decomposition-heavy pipeline, the local adaptivity of KAN may be weakened, so that the observed performance is driven more by the surrounding architecture than by KAN itself. This interpretation remains speculative, as we have not directly rerun TimeKAN's ablations under a unified protocol.

\noindent\textbf{Structural prior and multi-scale ablation}

We further examine how KAN interacts with explicit structural priors and the multi-scale design by introducing DFT and moving-average (MA) decomposition, and by removing the multi-scale module (NoMS). As shown in Fig.~\ref{fig2}(c,f), these modifications consistently hurt the KAN variant, while producing only marginal gains or mixed effects for the MLP counterpart. In particular, both decomposition-based priors lead to clear error increases for KAN, suggesting that explicitly imposed structure may interfere with the flexibility of the learned KAN representation. Removing the multi-scale module also degrades performance, indicating that KAN benefits from enriched multi-resolution inputs as long as they do not impose overly rigid structural constraints.

%\begin{table}[t]
%\centering
%\small
%\setlength{\tabcolsep}{3pt}
%\begin{tabular}{lccccc}
%\hline
%\multirow{2}{*}{Model} & \multicolumn{2}{c}{ETTh1} & \multicolumn{2}{c}{ETTm1} & \multirow{2}{*}{$\Delta$ MSE} \\
%                       & MSE & MAE & MSE & MAE & \\
%\hline
%MLP          & 0.459 & 0.445 & 0.392 & 0.411 & N/A \\
%MLP\_DFT     & 0.456 & 0.452 & 0.388 & 0.402 & -0.003 ($\downarrow$) \\
%MLP\_MA      & 0.448 & 0.447 & 0.381 & 0.405 & -0.011 ($\downarrow$) \\
%MLP\_NoMS    & 0.464 & 0.441 & 0.398 & 0.416 & +0.005 ($\uparrow$) \\
%\hline
%KAN          & \textbf{0.419} & \textbf{0.430} & \textbf{0.377} & \textbf{0.394} & N/A \\
%KAN\_DFT     & 0.444 & 0.447 & 0.387 & 0.401 & +0.025 ($\uparrow$) \\
%KAN\_MA      & 0.452 & 0.450 & 0.384 & 0.400 & +0.033 ($\uparrow$) \\
%KAN\_NoMS    & 0.439 & 0.438 & 0.383 & 0.397 & +0.020 ($\uparrow$) \\
%\hline
%\end{tabular}
%\caption{Structural prior ablation. Positive $\Delta$ MSE = degradation; negative = improvement.}
%\label{tab:ablation_decomposition}
%\end{table}

\subsection*{Computational efficiency}
\begin{figure*}[t]
\centering
\includegraphics[width=0.68\linewidth]{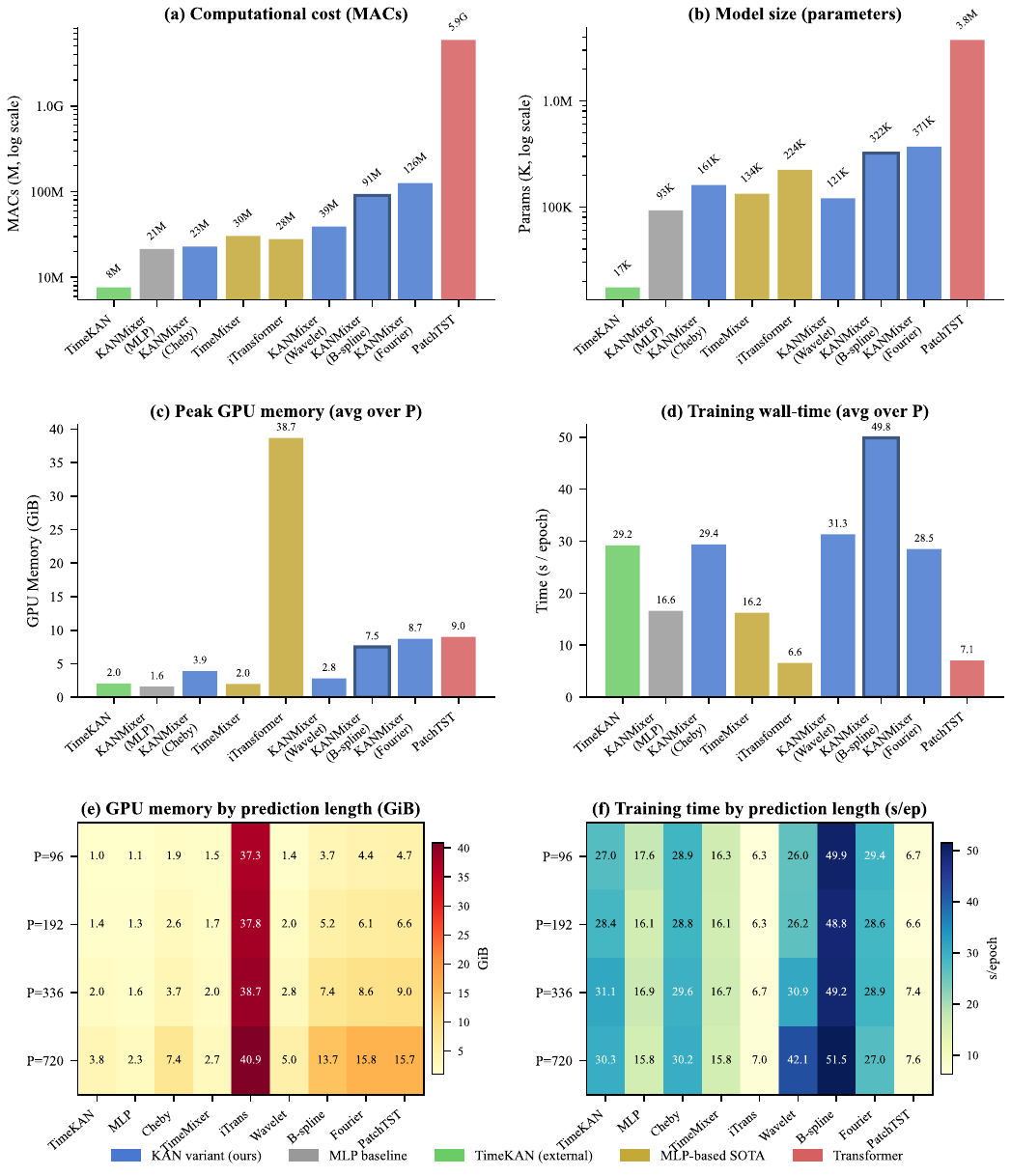}
\caption{Computational efficiency comparison on ETTh1. (a,b) MACs and parameter counts (both in log scale) for KANMixer variants and representative baselines. (c,d) Peak GPU memory and training wall-time averaged over prediction lengths $P \in \{96,192,336,720\}$. (e,f) GPU memory and training time broken down by prediction length. KANMixer remains parameter-efficient and substantially more memory-friendly than iTransformer, but KAN variants, especially the B-spline version, incur noticeably higher wall-clock cost than MLP baselines despite their relatively small model size.}
\label{fig3}
\end{figure*}

Fig.~\ref{fig3} summarizes the computational profile of KANMixer and several representative baselines. In terms of parameter count, KANMixer remains highly compact: the MLP baseline uses 92.9K parameters, while the Chebyshev, Wavelet, B-spline, and Fourier variants use 160.9K, 120.7K, 321.7K, and 371.1K parameters, respectively. All of these are substantially smaller than PatchTST, which requires 3.75M parameters. The arithmetic cost varies more noticeably across basis functions. Chebyshev KANMixer is computationally close to the MLP baseline, requiring 22.93M MACs versus 21.44M, whereas Wavelet, B-spline, and Fourier increase the cost to 39.12M, 90.57M, and 126.04M MACs, respectively. Even so, the most expensive KANMixer variant remains far lighter than PatchTST at 5.89G MACs.

GPU memory and wall-clock training reveal a different trade-off. Averaged across prediction lengths, the MLP baseline uses 1.6\,GiB of GPU memory, whereas B-spline and Fourier KANMixer require 7.5\,GiB and 8.7\,GiB, respectively; however, they remain far below iTransformer, whose memory footprint reaches 38.7\,GiB on average. This scaling trend becomes more pronounced at longer horizons: at $P{=}720$, B-spline and Fourier require 13.7\,GiB and 15.8\,GiB, compared with 2.3\,GiB for the MLP baseline and 40.9\,GiB for iTransformer. Training time shows an even clearer implementation bottleneck. Although Chebyshev KANMixer has nearly the same MAC count as the MLP baseline, it requires 29.4\,s/epoch on average, compared with 16.6\,s/epoch for MLP. The B-spline variant is the slowest in practice, reaching 49.8\,s/epoch on average and 51.5\,s/epoch at $P{=}720$, while PatchTST trains in only 7.1\,s/epoch on average despite its much larger model size. These results suggest that the current practical overhead of KANMixer is driven less by nominal MACs or parameter count than by implementation efficiency, especially the lack of optimized kernels for basis-function evaluation. Overall, KANMixer offers a favorable parameter-efficiency profile and moderate memory usage, but its wall-clock cost remains an engineering bottleneck.

\section*{Discussion}
% ============================================================

The five ablation studies together provide a consistent picture of how KAN behaves in LTSF. In our experiments, KAN is most effective when the basis function is chosen appropriately, when the architecture allows KAN to learn from enriched but unconstrained inputs, and when adaptive nonlinearity is preserved at the prediction head rather than replaced by a fixed-form MLP. Below, we discuss each of these findings in relation to prior literature.

\subsection*{KAN prediction head as the primary driver}
Component-wise ablation indicates that the KAN prediction head is the most influential module within KANMixer. Replacing it with an MLP while retaining KAN in the temporal mixing backbone leads to the largest accuracy drop, larger than replacing either KAN-FFN or KAN-Mixing individually. The final mapping from latent representations to $P$-step forecasts places the greatest demand on flexible function approximation, whereas the intermediate mixing layers operate on representations that have already been partially structured by residual connections and multi-scale aggregation. Under this view, adaptive B-spline functions may be especially useful at the prediction stage.

A practical implication is that introducing KAN only at the prediction head may be a relatively simple and efficient modification worth testing in existing LTSF architectures, rather than requiring a full architectural redesign. This perspective may also help explain TimeKAN's~\cite{huang2025timekan} finding that KAN and MLP behave similarly within its frequency decomposition pipeline: if most of the benefit arises at the prediction head, then placing KAN mainly in intermediate layers while keeping an MLP prediction head could attenuate the overall gain.

\subsection*{Basis function choice strongly influences KAN performance}
Basis function choice is one of the most important design variables in our experiments. B-spline functions provide local support and adaptive knot placement, which may help the model capture non-stationary temporal patterns, including localized fluctuations and abrupt changes, without globally altering the learned representation. This interpretation is consistent with the theoretical characterization of B-splines as locally supported piecewise polynomial functions, in which only a small number of basis elements are active at a given input value, enabling fine-grained adaptation~\cite{liu2025kan}. By contrast, Fourier and Wavelet bases impose more global or semi-global functional structure and, in our experiments, consistently underperform even a plain MLP. The Chebyshev basis is computationally more efficient~\cite{chebykan}, but it does not provide the same degree of local support as B-splines and degrades at longer prediction horizons on both ETTh1 and ETTm1. The theoretical connection between KAN and radial basis function networks~\cite{kan2rbf} offers a complementary perspective: the advantage of B-spline KAN may reflect the value of a more locally sensitive functional parameterization in temporally non-stationary settings.

Taken together, these findings suggest that basis function choice should be treated as a key design choice rather than a secondary implementation detail when applying KAN to temporal modeling. This contrasts with Chebyshev- and Fourier-based KAN variants, which have often been motivated primarily by computational efficiency~\cite{chebykan,kaf}.

\subsection*{Decomposition priors and KAN's adaptive learning}
One of the more unexpected findings in our experiments is that standard decomposition priors, implemented here as DFT and Moving Average decomposition, consistently degrade KAN performance (+0.025 to +0.033 MSE) while yielding small improvements for MLP ($-0.003$ to $-0.011$ MSE) under the same protocol. This pattern differs from prior LTSF studies, including DLinear~\cite{dlinear}, Autoformer~\cite{autoformer}, TimeMixer~\cite{wang2024timemixer}, and the theoretical analysis in~\cite{deng2024parsimony}, which generally report decomposition as beneficial for MLP-style architectures.

A possible explanation is that decomposition plays different roles for MLP and KAN. For MLP, whose nonlinearity is fixed in form, separating the input into trend and seasonal components may provide a useful inductive bias that simplifies the function to be learned by each branch. For KAN, by contrast, adaptive B-spline basis functions may already be flexible enough to model frequency and trend structure directly from raw inputs. Under this interpretation, imposing decomposition may restrict KAN to pre-filtered components and reduce the structured variation that its locally adaptive basis functions could otherwise exploit. This explanation remains speculative, and additional evidence, such as analyses of intermediate representations, would be needed to verify it.

Multi-scale pooling, the other form of structural enrichment tested here, shows the opposite pattern: removing it worsens KAN performance by about 0.020 MSE. This suggests that KAN may benefit from multi-resolution input enrichment when such enrichment preserves the original signal structure rather than explicitly decomposing it into predefined components. More broadly, this distinction between enrichment and decomposition may serve as a useful guideline when adapting existing LTSF design priors to KAN-based architectures.

\subsection*{Depth, stability, and KAN training dynamics}
A moderate stack depth ($N = 3$ KAN blocks) performs best across all three datasets, outperforming both shallower ($N = 2$) and deeper ($N = 4$) variants. The training instability observed for KAN-4L, including occasional divergence in fewer than 5\% of runs, is not observed for KAN-3L. This pattern is consistent with prior reports that KAN can be sensitive to initialization and gradient propagation in deeper stacks~\cite{tran2024}. It also differs from the behavior of MLP variants in our experiments, where performance changes more gradually across depths 2--4.

These results suggest that, in the current LTSF setting, increasing KAN depth beyond a moderate level may offer limited benefit while making optimization less stable. At present, the trade-off appears to favor relatively shallow or moderate-depth KAN stacks. Future work on initialization strategies, residual design~\cite{resnet}, or gradient-stabilizing training methods may help extend this range.

\subsection*{Positioning relative to concurrent KAN-for-LTSF work}
KANMixer differs from the two most closely related concurrent approaches~\cite{huang2025timekan,han2025kanseffectivemultivariatetime} in its emphasis on architectural simplicity. TimeKAN combines KAN with a frequency-decomposition pipeline, while RMoK places KAN within a mixture-of-experts framework. Both designs achieve competitive results, but they also introduce additional sources of variation that make it more difficult to isolate the contribution of KAN itself. By contrast, KANMixer is intentionally designed as a minimal testbed, allowing a more direct assessment of where KAN contributes within an LTSF architecture.

This design choice comes with a clear limitation. On high-dimensional multivariate datasets, architectures with explicit cross-variate modeling, such as iTransformer~\cite{liu2024itransformer}, or large-scale pre-training, such as Time-FFM~\cite{liu2024timeffm} and TimesFM~\cite{timesfm}, can provide advantages that a channel-independent model like KANMixer does not capture. In this sense, KANMixer should be viewed primarily as a diagnostic and reference architecture for studying KAN in LTSF, rather than as a production-oriented forecasting system.

\subsection*{Limitations and future directions}
Several limitations should be noted. First, the ablation findings reported here, including those on depth, component choice, basis function, decomposition, and multi-scale design, are established on three datasets (ETTh1, ETTm1, and Weather) under a single experimental protocol. These datasets cover hour- and minute-level forecasting in univariate and relatively low-dimensional multivariate settings, and the results should therefore not be assumed to generalize directly to strongly cross-correlated high-dimensional multivariate datasets, such as Electricity, to settings with periodic structure beyond the look-back window, to more severe distribution shifts than those handled by instance normalization, or to substantially longer prediction horizons.

Second, KANMixer adopts a channel-independent design and therefore does not explicitly model cross-variable dependencies. This places it at a disadvantage on datasets where cross-variate interactions are important. For example, on Electricity, architectures such as iTransformer~\cite{liu2024itransformer} can benefit from multivariate attention mechanisms that KANMixer does not include in its current form. Third, our explanations for why decomposition degrades KAN and why the prediction head contributes most are empirically motivated interpretations rather than formally established mechanisms, and would require additional targeted experiments to validate more directly.

A further limitation is computational. B-spline KAN incurs a substantial training-time overhead relative to MLP, approximately threefold per epoch in our experiments, largely because spline evaluation is not yet supported by comparably optimized CUDA kernels. In addition, the main comparison table combines three reproduced models with seven literature-reported results, so these two types of comparison should be interpreted differently.

Several directions for future work follow naturally from these limitations. One is to extend KANMixer with lightweight cross-variate modeling, inspired by architectures such as iTransformer~\cite{liu2024itransformer} and SOFTS~\cite{softs}. Others include evaluating KAN components under stronger distribution shift and concept drift, developing optimized spline kernels to reduce the computational gap, testing whether the prediction-head finding transfers to Transformer-based LTSF models, and using synthetic datasets with controlled non-stationarity to better characterize when local adaptivity is most beneficial.

\subsection*{Summary and broader implications}
Overall, our results provide empirical evidence that KAN components, when configured appropriately, can improve LTSF performance relative to MLP counterparts within the architectural settings studied here. The gains are most pronounced at the prediction head, where adaptive nonlinearity appears to be especially beneficial. A notable finding is that decomposition, which is often helpful in MLP-based LTSF models, degrades performance when combined with KAN in our experiments. This suggests that structural priors developed for MLP-based forecasting may not transfer directly to KAN-based architectures, and that the choice of backbone nonlinearity can influence which priors remain effective. In this sense, KANMixer serves as a reproducible reference architecture for studying KAN in LTSF, and our results underscore the value of controlled experiments that vary architectural structure and nonlinearity separately when assessing new forecasting components.

% ============================================================
\section*{Methods}
% ============================================================
\subsection*{Problem formulation}

Given a historical multivariate time series
\(
X = \{X_t^1, \ldots, X_t^d\}_{t=1}^{L}
\)
with $d$ variables and look-back window length $L$, the goal of long-term time series forecasting (LTSF) is to predict the next $P$ time steps,
\(
\hat{X} = f(X)
\).
Following the channel-independent protocol~\cite{patchtst}, each variable is forecast independently while sharing the same model parameters.

For a batch of input sequences, the model input has shape
\(
(B, L, C)
\),
where $B$ is the batch size, $L$ the input length, and $C$ the number of variables. The model output has shape
\(
(B, P, C)
\),
where $P$ is the prediction horizon. The primary training loss and evaluation metric is mean squared error (MSE):
\begin{equation}
\text{MSE} =
\frac{1}{P \times d}
\sum_{t=L+1}^{L+P}
\sum_{i=1}^{d}
\left(X_t^i - \hat{X}_t^i\right)^2.
\end{equation}
Mean absolute error (MAE) is reported as a secondary metric.

We use $k$ to denote the number of multi-scale levels, $N$ the number of temporal mixing blocks, $G$ the B-spline grid size, $p$ the spline order, $d_{\text{model}}$ the latent dimension, and $d_{\text{ff}}$ the hidden dimension used in the KAN-based feed-forward transformation. Intermediate hidden representations are denoted by $\mathbf{Z}_l^{(i)}$, where $l$ indexes the block and $i$ the scale. Unless otherwise stated, we use $G=5$ and $p=3$ by default.

\subsection*{Overview of KANMixer}

\begin{figure*}[t]
\centering
\includegraphics[width=0.48\linewidth]{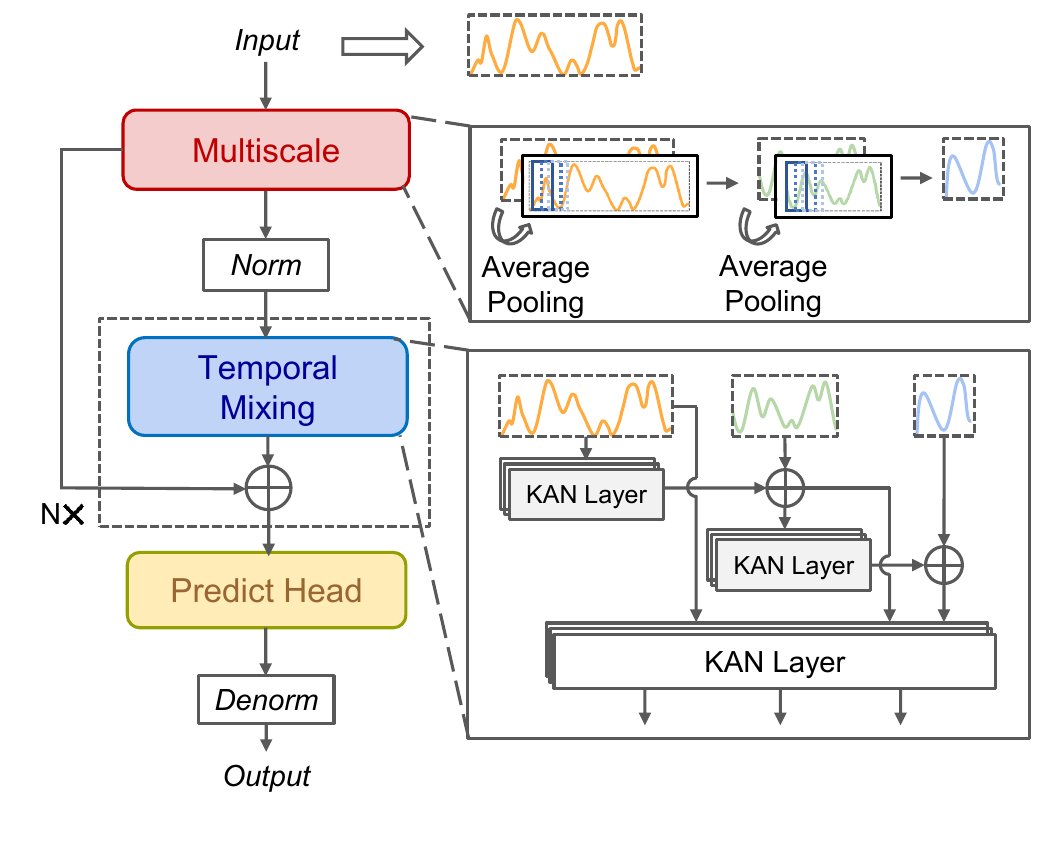}
\caption{Detailed architecture of KANMixer. The model consists of three stages: (1) multi-scale preprocessing and scale-wise embedding, (2) a KAN-based temporal mixing backbone with residual connections, and (3) independent scale-specific KAN prediction heads followed by additive aggregation.}
\label{fig:architecture}
\end{figure*}
KANMixer is a channel-independent multi-scale forecasting architecture built around a KAN-based temporal mixing backbone and scale-specific KAN prediction heads, as illustrated in Fig.~\ref{fig:architecture}. Given an input sequence of shape
\(
(B, L, C)
\),
KANMixer first constructs a list of downsampled sequences at multiple temporal resolutions. Each scale is normalized independently and embedded into a latent representation of shape
\(
(B C, L_i, d_{\text{model}})
\),
where $L_i$ is the temporal length at scale $i$. These scale-specific latent sequences are then processed jointly by a stack of $N$ temporal mixing blocks. Finally, each scale produces its own forecast through scale-specific KAN prediction heads, and the forecasts are summed across scales before inverse normalization.

\subsection*{Multi-scale preprocessing and embedding}

For an input tensor
\(
\mathbf{X} \in \mathbb{R}^{B \times L \times C}
\),
KANMixer first constructs a set of $k$ temporal scales using repeated pooling over the time dimension. Let
\(
\mathbf{X}^{(0)}, \mathbf{X}^{(1)}, \ldots, \mathbf{X}^{(k-1)}
\)
denote the resulting multi-scale sequence list, where
\(
\mathbf{X}^{(i)} \in \mathbb{R}^{B \times L_i \times C}
\)
and $L_0=L$. In our default setting, three scales are used. With a downsampling window of 2, the corresponding temporal lengths are approximately $L$, $L/2$, and $L/4$.

Each scale is normalized independently using reversible instance normalization (RevIN)~\cite{kim2022revin}. For variable $c$ at scale $i$, the normalized sequence is
\begin{equation}
\tilde{X}_{t,c}^{(i)}=
\frac{X_{t,c}^{(i)}-\mu_c^{(i)}}
{\sigma_c^{(i)}+\epsilon},
\end{equation}
where
\begin{equation}
\mu_c^{(i)}=
\frac{1}{L_i}
\sum_{t=1}^{L_i} X_{t,c}^{(i)},
\qquad
\sigma_c^{(i)}=
\sqrt{
\frac{1}{L_i}
\sum_{t=1}^{L_i}
\left(X_{t,c}^{(i)}-\mu_c^{(i)}\right)^2
}.
\end{equation}
Normalization is applied separately to each scale before embedding. After scale-wise forecasting and additive aggregation, inverse normalization is performed using the RevIN statistics from the finest (full-resolution) scale. In the controlled reproduced models, the same RevIN configuration is applied unless otherwise stated.

Under the channel-independent protocol, each variable is treated as an independent sequence. Therefore, for each scale $i$, the normalized tensor
\(
\tilde{\mathbf{X}}^{(i)} \in \mathbb{R}^{B \times L_i \times C}
\)
is reshaped into
\(
\mathbb{R}^{(B C) \times L_i \times 1}
\),
so that each channel becomes an individual sample. A shared embedding layer then maps each univariate sequence into a latent representation:
\begin{equation}
\mathbf{E}^{(i)} = \text{Embed}\!\left(\tilde{\mathbf{X}}^{(i)}\right),
\qquad
\mathbf{E}^{(i)} \in \mathbb{R}^{(B C) \times L_i \times d_{\text{model}}}.
\end{equation}
This embedding is implemented by a position-free token embedding layer (`DataEmbedding\_wo\_pos`), which projects the scalar input at each time step into a $d_{\text{model}}$-dimensional latent space.

As a result, the frontend outputs a list of $k$ scale-specific latent sequences,
\(
\{\mathbf{E}^{(0)}, \ldots, \mathbf{E}^{(k-1)}\}
\),
rather than collapsing all scales into a single tensor. This list is passed directly to the temporal mixing backbone.

\subsection*{Temporal mixing backbone}

The temporal mixing backbone consists of a stack of $N$ blocks. Each block takes as input a list of scale-specific latent tensors
\(
\{\mathbf{Z}_{l-1}^{(0)}, \ldots, \mathbf{Z}_{l-1}^{(k-1)}\}
\),
where
\(
\mathbf{Z}_{l-1}^{(i)} \in \mathbb{R}^{(B C) \times L_i \times d_{\text{model}}}
\),
and returns a list of the same form.

Each block performs two operations: (1) fine-to-coarse temporal mixing across scales, and (2) within-scale feature refinement.

\paragraph{Fine-to-coarse temporal mixing.}
For temporal mixing, each scale tensor is first transposed to
\(
\mathbb{R}^{(B C) \times d_{\text{model}} \times L_i}
\),
so that KAN can operate on the temporal dimension. For each adjacent pair of scales $(i, i+1)$, the finer-scale representation is projected from temporal length $L_i$ to $L_{i+1}$ using a KAN mapping:
\begin{equation}
\mathbf{M}_{l}^{(i+1)}
=
\mathbf{Z}_{l-1}^{(i+1)}
+
\text{KAN}_{\text{down}}^{(i)}
\!\left(\mathbf{Z}_{l-1}^{(i)}\right),
\qquad i=0,\ldots,k-2,
\end{equation}
where
\(
\text{KAN}_{\text{down}}^{(i)}:\mathbb{R}^{L_i}\rightarrow\mathbb{R}^{L_{i+1}}
\)
is applied independently to each latent feature channel. In practice, each down-mixing module is implemented as a small stack of KAN layers, with the first layer reducing the temporal length and the subsequent layers refining the projected representation at the lower scale. The finest scale is preserved directly and serves as the starting point of the cascade.

This operation propagates information from finer to coarser temporal resolutions while preserving scale-specific streams.

\paragraph{Within-scale feature refinement.}
After temporal mixing, each scale is transposed back to
\(
\mathbb{R}^{(B C) \times L_i \times d_{\text{model}}}
\).
A KAN-based fusion module is then applied along the feature dimension:
\begin{equation}
\mathbf{Z}_{l}^{(i)}
=
\mathbf{M}_{l}^{(i)}
+
\text{KAN}_{\text{ffn}}^{(i)}
\!\left(\mathbf{M}_{l}^{(i)}\right),
\qquad i=0,\ldots,k-1,
\end{equation}
where
\(
\text{KAN}_{\text{ffn}}^{(i)}:\mathbb{R}^{d_{\text{model}}}\rightarrow\mathbb{R}^{d_{\text{model}}}
\)
is implemented as a residual KAN feed-forward transformation. Thus, each block combines cross-scale temporal interaction with within-scale nonlinear refinement. In our implementation, this feed-forward transformation uses an internal hidden width $d_{\text{ff}}$; unless otherwise stated, we set $d_{\text{ff}}=32$.

All KAN layers in the backbone use B-spline basis functions with grid size $G=5$ and spline order $p=3$ unless otherwise stated. Residual connections are applied throughout to improve optimization stability. After $N$ blocks, the backbone outputs
\(
\{\mathbf{Z}_{N}^{(0)}, \ldots, \mathbf{Z}_{N}^{(k-1)}\}
\),
with the same shapes as the input list.

\subsection*{Prediction heads and aggregation}

KANMixer uses an independent prediction head for each scale. For scale $i$, the final backbone output
\(
\mathbf{Z}_{N}^{(i)} \in \mathbb{R}^{(B C) \times L_i \times d_{\text{model}}}
\)
is first transposed to
\(
\mathbb{R}^{(B C) \times d_{\text{model}} \times L_i}
\).
A scale-specific KAN prediction layer then maps the temporal dimension from $L_i$ to the prediction horizon $P$:
\begin{equation}
\mathbf{Q}^{(i)}
=
\text{KAN}_{\text{pred}}^{(i)}
\!\left(\mathbf{Z}_{N}^{(i)}\right),
\qquad
\mathbf{Q}^{(i)} \in \mathbb{R}^{(B C) \times d_{\text{model}} \times P}.
\end{equation}
This operation produces a scale-specific prediction over the horizon for each latent feature channel.

The result is then transposed back to
\(
\mathbb{R}^{(B C) \times P \times d_{\text{model}}}
\),
and a final KAN output projection maps the feature dimension from $d_{\text{model}}$ to 1:
\begin{equation}
\hat{\mathbf{Y}}^{(i)}
=
\text{KAN}_{\text{out}}
\!\left(\mathbf{Q}^{(i)}\right),
\qquad
\hat{\mathbf{Y}}^{(i)} \in \mathbb{R}^{(B C) \times P \times 1}.
\end{equation}
After reshaping, each scale yields a forecast tensor of shape
\(
\mathbb{R}^{B \times P \times C}
\).

The final prediction is obtained by stacking the scale-specific forecasts and summing over the scale dimension:
\begin{equation}
\hat{\mathbf{Y}}
=
\sum_{i=0}^{k-1}
\hat{\mathbf{Y}}^{(i)},
\qquad
\hat{\mathbf{Y}} \in \mathbb{R}^{B \times P \times C}.
\end{equation}
No learnable weighting is used in this aggregation step, which keeps the architecture simple and avoids introducing additional fusion parameters.

Finally, inverse normalization is applied using the RevIN statistics of the finest (full-resolution) scale, yielding the final output in the original data space. The complete forward pass therefore maps
\(
(B,L,C)\rightarrow(B,P,C)
\)
through scale construction, scale-wise embedding, KAN-based temporal mixing, scale-wise prediction, additive aggregation, and denormalization.

\noindent\textbf{KAN layer parameterization.}
For the B-spline variant used in KANMixer, each KAN layer maps an input vector
$\mathbf{x}\in\mathbb{R}^{n_{\text{in}}}$ to an output
$\mathbf{y}\in\mathbb{R}^{n_{\text{out}}}$ through learnable univariate edge functions:
\begin{equation}
y_j=\sum_{i=1}^{n_{\text{in}}}\phi_{j,i}(x_i).
\end{equation}
In our implementation, each edge function is parameterized as the sum of a base branch and a spline branch,
\begin{equation}
\phi_{j,i}(x)
=
w^{(b)}_{j,i}\,\sigma(x)
+
\sum_{g=1}^{G+p}\tilde{c}_{j,i,g}\,B_{g,p}(x),
\end{equation}
where $\sigma(\cdot)$ is the SiLU activation, $B_{g,p}$ denotes the $g$-th B-spline basis function of order $p$, and $\tilde{c}_{j,i,g}$ are learnable spline coefficients after scaling. Unless otherwise stated, we use grid size $G=5$ and spline order $p=3$. The B-spline grid is initialized uniformly over $[-1,1]$ and extended by $p$ boundary knots on both sides, following the standard spline construction. Compared with an MLP layer, the KAN layer therefore includes both a base linear branch and a spline branch, which increases parameter count and computation.

\noindent\textbf{Architecture hyperparameters.} The latent dimension is $d_{\text{model}}=16$; the temporal mixing backbone uses $N=3$ KAN blocks (established by depth ablation). Multi-scale levels: $k=3$ with average pooling kernel sizes $\{1, 2, 4\}$. Total parameter count at default settings: 321.73K (B-spline KANMixer), 92.9K (MLP counterpart). Model width is tuned for KAN and MLP variants independently via grid search over $d_{\text{model}} \in \{16,32, 64, 128\}$; KAN achieves peak performance at $d_{\text{model}}=16$, while MLP variants show weaker width sensitivity.

\noindent\textbf{Training stability protocol.} Gradient clipping at max norm 1.0 is applied uniformly. A divergence criterion is applied to all reproduced models: if the training loss at any epoch exceeds $10\times$ the epoch-1 loss, the run is discarded and restarted with a new random seed (seed offset by 1). This criterion was triggered in fewer than 5\% of KAN-4L runs and was not observed for KAN-3L or any MLP variant. Five-run averages reflect five converged runs. Early stopping with a patience of 3 epochs based on validation MSE is applied, with training capped at 10 total epochs.

\noindent\textbf{Software and hardware.} Reversible instance normalization (RevIN~\cite{kim2022revin}) is applied at input and inverse-applied at output, using per-channel mean and variance estimated on each training window. All experiments used PyTorch 2.1 with CUDA 12.1 on 4$\times$ NVIDIA A100 80G GPUs.

\subsection*{Baselines and evaluation protocol}

We strictly follow the established LTSF evaluation protocol of~\cite{autoformer}: Adam optimizer with learning rate 0.001 and cosine annealing decay, batch size 32, MSE training loss, and standardized data splits. ETT datasets (ETTh1, ETTh2, ETTm1, ETTm2) use 6:2:2 train/validation/test splits; Exchange Rate, Weather, and Electricity use 7:1:2 splits, following~\cite{autoformer}.

\textbf{Reproduced models}: KANMixer, TimeKAN~\cite{huang2025timekan}, and TimeMixer~\cite{wang2024timemixer} were implemented and trained in our environment (4$\times$ NVIDIA A100 80G GPUs, Tohoku University HPC cluster) using each model's official released code with default hyperparameters, except that RevIN is applied consistently to all three. Results are averaged over five independent runs with different random seeds; standard deviations are $\leq 0.005$ MSE and $\leq 0.003$ MAE across all 28 benchmark--horizon settings.

\textbf{Literature-reported baselines}: iTransformer~\cite{liu2024itransformer}, PatchTST~\cite{patchtst}, Time-FFM~\cite{liu2024timeffm}, TimesNet~\cite{wu2023timesnet}, FreTS~\cite{yi2023frequencydomain}, DLinear~\cite{dlinear}, and TiDE~\cite{tide} use single-run numbers from their respective publications. Comparison with these baselines should be treated as contextual rather than strictly controlled, as differences in GPU hardware, software environment, random seed, and the presence or absence of RevIN across original papers may introduce variability beyond model-intrinsic performance.

\textbf{Hyperparameter sensitivity.} The main results use the default configuration ($N=3$, $k=3$, $d_{\text{model}}=16$, $d_{\text{ff}}=32$, $G=5$, $p=3$), selected based on validation MSE on ETTh1 and then fixed for all test-set evaluations. Sensitivity to $N$ is examined through the depth ablation. Sensitivity to $d_{\text{model}}$ is evaluated via grid search over $\{16, 32, 64, 128\}$. The grid size $G$ and polynomial order $p$ follow the default settings in~\cite{liu2025kan} and are not independently ablated here. The number of pooling scales is fixed to $k=3$ to match TimeMixer~\cite{wang2024timemixer} for comparability, and its sensitivity is not separately evaluated.

\textbf{Computational cost measurements.} MACs, parameter counts, peak GPU memory, and per-epoch wall-time are measured on a single NVIDIA A100 80G GPU with batch size 32 and look-back window $L=96$, using ETTh1 as a representative benchmark. Inference time (not reported) scales approximately linearly with prediction length $P$.

% ============================================================
% END MATTER
% ============================================================

\section*{Data availability}

All datasets used in this study are publicly available benchmark datasets. 
The ETT series (ETTh1, ETTh2, ETTm1, ETTm2), Weather, Electricity, and Exchange Rate datasets 
are accessible from the official Time-Series-Library repository at 
https://github.com/thuml/Time-Series-Library.

No new datasets were generated in this study. 
The code and processed data required to reproduce the results 
will be made available to editors and reviewers upon request 
and will be publicly released upon acceptance.

\section*{Code availability}
The complete implementation of KANMixer, including model definitions, training scripts, evaluation code, hyperparameter configuration files, and environment specification, will be provided as a supplementary code archive to editors and reviewers upon initial submission. The code and all scripts needed to reproduce the results from the hyperparameters specified in the Methods section will be made publicly available upon acceptance.

\section*{Funding}
The authors received no specific funding for this work.

\section*{Author contributions}
L.J. conceived the architecture and led the experimental design. D.H. performed the ablation studies (basis function comparison, structural prior analysis, computational efficiency analysis, component-wise ablation, KAN-vs-MLP depth analysis) and contributed to writing and revision. Y.W., Y.S., S.X., W.C., X.Z., Z.T., and Z.Z. contributed to methodology discussion, experimental evaluation, and manuscript review. F.L. supervised the project and directed the research. M.Z. and K.Y. provided guidance on architecture design and manuscript revision.

\section*{Competing interests}
The authors declare no competing interests.

\bibliography{ref}

@article{ltsf,
title = {Long sequence time-series forecasting with deep learning: A survey},
journal = {Information Fusion},
volume = {97},
pages = {101819},
year = {2023},
issn = {1566-2535},
doi = {https://doi.org/10.1016/j.inffus.2023.101819},
url = {https://www.sciencedirect.com/science/article/pii/S1566253523001355},
author = {Zonglei Chen and Minbo Ma and Tianrui Li and Hongjun Wang and Chongshou Li},
}

@article{ke2017lightgbm,
  title={Lightgbm: A highly efficient gradient boosting decision tree},
  author={Ke, Guolin and Meng, Qi and Finley, Thomas and Wang, Taifeng and Chen, Wei and Ma, Weidong and Ye, Qiwei and Liu, Tie-Yan},
  journal={Advances in neural information processing systems},
  volume={30},
  year={2017}
}

@Article{en13246623,
AUTHOR = {Rajagukguk, Rial A. and Ramadhan, Raden A. A. and Lee, Hyun-Jin},
TITLE = {A Review on Deep Learning Models for Forecasting Time Series Data of Solar Irradiance and Photovoltaic Power},
JOURNAL = {Energies},
VOLUME = {13},
YEAR = {2020},
NUMBER = {24},
ARTICLE-NUMBER = {6623},
URL = {https://www.mdpi.com/1996-1073/13/24/6623},
ISSN = {1996-1073},
DOI = {10.3390/en13246623}
}

@misc{traffic,
      title={Virtual Nodes Improve Long-term Traffic Prediction},
      author={Xiaoyang Cao and Dingyi Zhuang and Jinhua Zhao and Shenhao Wang},
      year={2025},
      eprint={2501.10048},
      archivePrefix={arXiv},
      primaryClass={cs.LG},
      url={https://arxiv.org/abs/2501.10048},
}

@book{mse,
title = "Forecasting: Principles and Practice",
author = "Hyndman, {Robin John} and George Athanasopoulos",
year = "2018",
language = "English",
publisher = "OTexts",
address = "Australia",
edition = "2nd",
}

@inproceedings{
liu2024timeffm,
title={Time-{FFM}: Towards {LM}-Empowered Federated Foundation Model for Time Series Forecasting},
author={Qingxiang Liu and Xu Liu and Chenghao Liu and Qingsong Wen and Yuxuan Liang},
booktitle={The Thirty-eighth Annual Conference on Neural Information Processing Systems},
year={2024},
url={https://openreview.net/forum?id=HS0faHRhWD}
}

@article{
tide,
title={Long-term Forecasting with Ti{DE}: Time-series Dense Encoder},
author={Abhimanyu Das and Weihao Kong and Andrew Leach and Shaan K Mathur and Rajat Sen and Rose Yu},
journal={Transactions on Machine Learning Research},
issn={2835-8856},
year={2023},
url={https://openreview.net/forum?id=pCbC3aQB5W},
note={}
}

@article{vaswani2017attention,
  title={Attention is all you need},
  author={Vaswani, Ashish and Shazeer, Noam and Parmar, Niki and Uszkoreit, Jakob and Jones, Llion and Gomez, Aidan N and Kaiser, {\L}ukasz and Polosukhin, Illia},
  journal={Advances in neural information processing systems},
  volume={30},
  year={2017}
}

@inproceedings{
liu2024itransformer,
title={iTransformer: Inverted Transformers Are Effective for Time Series Forecasting},
author={Yong Liu and Tengge Hu and Haoran Zhang and Haixu Wu and Shiyu Wang and Lintao Ma and Mingsheng Long},
booktitle={The Twelfth International Conference on Learning Representations},
year={2024},
url={https://openreview.net/forum?id=JePfAI8fah}
}

@misc{position,
      title={Position: There are no Champions in Long-Term Time Series Forecasting},
      author={Lorenzo Brigato and Rafael Morand and Knut Str{\o}mmen and Maria Panagiotou and Markus Schmidt and Stavroula Mougiakakou},
      year={2025},
      eprint={2502.14045},
      archivePrefix={arXiv},
      primaryClass={cs.LG},
      url={https://arxiv.org/abs/2502.14045},
}

@misc{electricity,
  author       = {Trindade, Artur},
  title        = {{ElectricityLoadDiagrams20112014}},
  year         = {2015},
  howpublished = {UCI Machine Learning Repository},
  note         = {{DOI}: https://doi.org/10.24432/C58C86}
}

@article{arima,
title = {The use of ARIMA models for reliability forecasting and analysis},
journal = {Computers \& Industrial Engineering},
volume = {35},
number = {1},
pages = {213-216},
year = {1998},
issn = {0360-8352},
doi = {https://doi.org/10.1016/S0360-8352(98)00066-7},
url = {https://www.sciencedirect.com/science/article/pii/S0360835298000667},
author = {S.L. Ho and M. Xie},
}

@ARTICLE{lstm,
  author={Hochreiter, Sepp and Schmidhuber, J{\"u}rgen},
  journal={Neural Computation},
  title={Long Short-Term Memory},
  year={1997},
  volume={9},
  number={8},
  pages={1735-1780},
  doi={10.1162/neco.1997.9.8.1735}}

@inproceedings{xgboost,
author = {Chen, Tianqi and Guestrin, Carlos},
title = {XGBoost: A Scalable Tree Boosting System},
year = {2016},
isbn = {9781450342322},
publisher = {Association for Computing Machinery},
address = {New York, NY, USA},
url = {https://doi.org/10.1145/2939672.2939785},
doi = {10.1145/2939672.2939785},
booktitle = {Proceedings of the 22nd ACM SIGKDD International Conference on Knowledge Discovery and Data Mining},
pages = {785--794},
numpages = {10},
location = {San Francisco, California, USA},
series = {KDD '16}
}

@article{gnn1,
title = {Uncertainty-aware probabilistic graph neural networks for road-level traffic crash prediction},
journal = {Accident Analysis \& Prevention},
volume = {208},
pages = {107801},
year = {2024},
issn = {0001-4575},
doi = {https://doi.org/10.1016/j.aap.2024.107801},
url = {https://www.sciencedirect.com/science/article/pii/S0001457524003464},
author = {Xiaowei Gao and Xinke Jiang and James Haworth and Dingyi Zhuang and Shenhao Wang and Huanfa Chen and Stephen Law},
}

@misc{resnet,
      title={Deep Residual Learning for Image Recognition},
      author={Kaiming He and Xiangyu Zhang and Shaoqing Ren and Jian Sun},
      year={2015},
      eprint={1512.03385},
      archivePrefix={arXiv},
      primaryClass={cs.CV},
      url={https://arxiv.org/abs/1512.03385},
}

@inproceedings{
fredf,
title={Fre{DF}: Learning to Forecast in the Frequency Domain},
author={Hao Wang and Lichen Pan and Yuan Shen and Zhichao Chen and Degui Yang and Yifei Yang and Sen Zhang and Xinggao Liu and Haoxuan Li and Dacheng Tao},
booktitle={The Thirteenth International Conference on Learning Representations},
year={2025},
url={https://openreview.net/forum?id=4A9IdSa1ul}
}

@misc{
patchmixer,
title={PatchMixer: A Patch-Mixing Architecture for Long-Term Time Series Forecasting},
author={Zeying Gong and Yujin Tang and Junwei Liang},
year={2024},
url={https://openreview.net/forum?id=Te5v4EcFGL}
}

@article{kainn,
   title={Kolmogorov--Arnold-Informed neural network: A physics-informed deep learning framework for solving forward and inverse problems based on Kolmogorov--Arnold Networks},
   volume={433},
   ISSN={0045-7825},
   url={http://dx.doi.org/10.1016/j.cma.2024.117518},
   DOI={10.1016/j.cma.2024.117518},
   journal={Computer Methods in Applied Mechanics and Engineering},
   publisher={Elsevier BV},
   author={Wang, Yizheng and Sun, Jia and Bai, Jinshuai and Anitescu, Cosmin and Eshaghi, Mohammad Sadegh and Zhuang, Xiaoying and Rabczuk, Timon and Liu, Yinghua},
   year={2025},
   month=jan, pages={117518} }

@inproceedings{
huang2025timekan,
title={Time{KAN}: {KAN}-based Frequency Decomposition Learning Architecture for Long-term Time Series Forecasting},
author={Songtao Huang and Zhen Zhao and Can Li and LEI BAI},
booktitle={The Thirteenth International Conference on Learning Representations},
year={2025},
url={https://openreview.net/forum?id=wTLc79YNbh}
}

@article{weather,
  author={Haixu Wu and Hang Zhou and Mingsheng Long and Jianmin Wang},
  title={Interpretable weather forecasting for worldwide stations with a unified deep model},
  year={2023},
  month={June},
  cdate={1685577600000},
  journal={Nat. Mac. Intell.},
  volume={5},
  number={6},
  pages={602-611},
  url={https://doi.org/10.1038/s42256-023-00667-9}
}

@inproceedings{
autoformer,
title={Autoformer: Decomposition Transformers with Auto-Correlation for Long-Term Series Forecasting},
author={Haixu Wu and Jiehui Xu and Jianmin Wang and Mingsheng Long},
booktitle={Advances in Neural Information Processing Systems},
editor={A. Beygelzimer and Y. Dauphin and P. Liang and J. Wortman Vaughan},
year={2021},
url={https://openreview.net/forum?id=J4gRj6d5Qm}
}

@inproceedings{
zhang2023crossformer,
title={Crossformer: Transformer Utilizing Cross-Dimension Dependency for Multivariate Time Series Forecasting},
author={Yunhao Zhang and Junchi Yan},
booktitle={The Eleventh International Conference on Learning Representations},
year={2023},
url={https://openreview.net/forum?id=vSVLM2j9eie}
}

@misc{tsmixer,
      title={TSMixer: An All-MLP Architecture for Time Series Forecasting},
      author={Si-An Chen and Chun-Liang Li and Nate Yoder and Sercan O. Arik and Tomas Pfister},
      year={2023},
      eprint={2303.06053},
      archivePrefix={arXiv},
      primaryClass={cs.LG},
      url={https://arxiv.org/abs/2303.06053},
}

@inproceedings{
patchtst,
title={A Time Series is Worth 64 Words:  Long-term Forecasting with Transformers},
author={Yuqi Nie and Nam H Nguyen and Phanwadee Sinthong and Jayant Kalagnanam},
booktitle={The Eleventh International Conference on Learning Representations},
year={2023},
url={https://openreview.net/forum?id=Jbdc0vTOcol}
}

@article{kart,
title = {The Kolmogorov--Arnold representation theorem revisited},
journal = {Neural Networks},
volume = {137},
pages = {119-126},
year = {2021},
issn = {0893-6080},
doi = {https://doi.org/10.1016/j.neunet.2021.01.020},
url = {https://www.sciencedirect.com/science/article/pii/S0893608021000289},
author = {Johannes Schmidt-Hieber},
}

@misc{wavkan,
      title={Wav-KAN: Wavelet Kolmogorov-Arnold Networks},
      author={Zavareh Bozorgasl and Hao Chen},
      year={2024},
      eprint={2405.12832},
      archivePrefix={arXiv},
      primaryClass={cs.LG},
      url={https://arxiv.org/abs/2405.12832},
}

@misc{kaf,
      title={Kolmogorov-Arnold Fourier Networks},
      author={Jusheng Zhang and Yijia Fan and Kaitong Cai and Keze Wang},
      year={2025},
      eprint={2502.06018},
      archivePrefix={arXiv},
      primaryClass={cs.LG},
      url={https://arxiv.org/abs/2502.06018},
}

@misc{chebykan,
      title={Chebyshev Polynomial-Based Kolmogorov-Arnold Networks: An Efficient Architecture for Nonlinear Function Approximation},
      author={Sidharth SS and Keerthana AR and Gokul R and Anas KP},
      year={2024},
      eprint={2405.07200},
      archivePrefix={arXiv},
      primaryClass={cs.LG},
      url={https://arxiv.org/abs/2405.07200},
}

@misc{han2025kanseffectivemultivariatetime,
      title={Are KANs Effective for Multivariate Time Series Forecasting?},
      author={Xiao Han and Xinfeng Zhang and Yiling Wu and Zhenduo Zhang and Zhe Wu},
      year={2025},
      eprint={2408.11306},
      archivePrefix={arXiv},
      primaryClass={cs.LG},
      url={https://arxiv.org/abs/2408.11306},
}

@inproceedings{
liu2025kan,
title={{KAN}: Kolmogorov{\textendash}Arnold Networks},
author={Ziming Liu and Yixuan Wang and Sachin Vaidya and Fabian Ruehle and James Halverson and Marin Soljacic and Thomas Y. Hou and Max Tegmark},
booktitle={The Thirteenth International Conference on Learning Representations},
year={2025},
url={https://openreview.net/forum?id=Ozo7qJ5vZi}
}

@inproceedings{
wang2024timemixer,
title={TimeMixer: Decomposable Multiscale Mixing for Time Series Forecasting},
author={Shiyu Wang and Haixu Wu and Xiaoming Shi and Tengge Hu and Huakun Luo and Lintao Ma and James Y. Zhang and JUN ZHOU},
booktitle={The Twelfth International Conference on Learning Representations},
year={2024},
url={https://openreview.net/forum?id=7oLshfEIC2}
}

@inproceedings{wu2023timesnet,
  title={TimesNet: Temporal 2D-Variation Modeling for General Time Series Analysis},
  author={Haixu Wu and Tengge Hu and Yong Liu and Hang Zhou and Jianmin Wang and Mingsheng Long},
  booktitle={International Conference on Learning Representations},
  year={2023},
}

@inproceedings{
yi2023frequencydomain,
title={Frequency-domain {MLP}s are More Effective Learners in Time Series Forecasting},
author={Kun Yi and Qi Zhang and Wei Fan and Shoujin Wang and Pengyang Wang and Hui He and Ning An and Defu Lian and Longbing Cao and Zhendong Niu},
booktitle={Thirty-seventh Conference on Neural Information Processing Systems},
year={2023},
url={https://openreview.net/forum?id=iif9mGCTfy}
}

@article{dlinear, title={Are Transformers Effective for Time Series Forecasting?}, volume={37}, url={https://ojs.aaai.org/index.php/AAAI/article/view/26317}, DOI={10.1609/aaai.v37i9.26317}, number={9}, journal={Proceedings of the AAAI Conference on Artificial Intelligence}, author={Zeng, Ailing and Chen, Muxi and Zhang, Lei and Xu, Qiang}, year={2023}, month={Jun.}, pages={11121-11128} }

@InProceedings{fedformer,
  title = 	 {{FED}former: Frequency Enhanced Decomposed Transformer for Long-term Series Forecasting},
  author =       {Zhou, Tian and Ma, Ziqing and Wen, Qingsong and Wang, Xue and Sun, Liang and Jin, Rong},
  booktitle = 	 {Proceedings of the 39th International Conference on Machine Learning},
  pages = 	 {27268--27286},
  year = 	 {2022},
  editor = 	 {Chaudhuri, Kamalika and Jegelka, Stefanie and Song, Le and Szepesvari, Csaba and Niu, Gang and Sabato, Sivan},
  volume = 	 {162},
  series = 	 {Proceedings of Machine Learning Research},
  month = 	 {17--23 Jul},
  publisher =    {PMLR},
  url = 	 {https://proceedings.mlr.press/v162/zhou22g.html},
}

@misc{tran2024,
      title={Exploring the Limitations of Kolmogorov-Arnold Networks in Classification: Insights to Software Training and Hardware Implementation},
      author={Van Duy Tran and Tran Xuan Hieu Le and Thi Diem Tran and Hoai Luan Pham and Vu Trung Duong Le and Tuan Hai Vu and Van Tinh Nguyen and Yasuhiko Nakashima},
      year={2024},
      eprint={2407.17790},
      archivePrefix={arXiv},
      primaryClass={cs.LG},
      url={https://arxiv.org/abs/2407.17790},
}

@article{informer, title={Informer: Beyond Efficient Transformer for Long Sequence Time-Series Forecasting}, volume={35}, url={https://ojs.aaai.org/index.php/AAAI/article/view/17325}, DOI={10.1609/aaai.v35i12.17325}, number={12}, journal={Proceedings of the AAAI Conference on Artificial Intelligence}, author={Zhou, Haoyi and Zhang, Shanghang and Peng, Jieqi and Zhang, Shuai and Li, Jianxin and Xiong, Hui and Zhang, Wancai}, year={2021}, month={May}, pages={11106-11115} }

@article{d2021developing,
  title={Developing a novel recurrent neural network architecture with fewer parameters and good learning performance},
  author={D YAMADA, Kazunori and Lin, Fangzhou and Nakamura, Tsukasa},
  journal={Interdisciplinary information sciences},
  volume={27},
  number={1},
  pages={25--40},
  year={2021},
  publisher={The Editorial Committee of the Interdisciplinary Information Sciences}
}

@article{deng2024parsimony,
  title={Parsimony or capability? decomposition delivers both in long-term time series forecasting},
  author={Deng, Jinliang and Ye, Feiyang and Yin, Du and Song, Xuan and Tsang, Ivor and Xiong, Hui},
  journal={Advances in Neural Information Processing Systems},
  volume={37},
  pages={66687--66712},
  year={2024}
}

@inproceedings{jiang2025time,
  title={Time series supplier allocation via deep black-litterman model},
  author={Jiang, Xinke and Zhang, Wentao and Fang, Yuchen and Gao, Xiaowei and Chen, Hao and Zhang, Haoyu and Zhuang, Dingyi and Luo, Jiayuan},
  booktitle={Proceedings of the AAAI Conference on Artificial Intelligence},
  volume={39},
  pages={11870--11878},
  year={2025}
}

@inproceedings{wu2021inductive,
  title={Inductive graph neural networks for spatiotemporal kriging},
  author={Wu, Yuankai and Zhuang, Dingyi and Labbe, Aurelie and Sun, Lijun},
  booktitle={Proceedings of the AAAI conference on artificial intelligence},
  volume={35},
  pages={4478--4485},
  year={2021}
}

@inproceedings{kim2022revin,
title={Reversible Instance Normalization for Accurate Time-Series Forecasting against Distribution Shift},
author={Taesung Kim and Jinhee Kim and Yunwon Tae and Cheonbok Park and Jang-Ho Chun and Jaegul Choo},
booktitle={International Conference on Learning Representations},
year={2022},
url={https://openreview.net/forum?id=cGDAkQo1C0p}
}

@misc{kan2rbf,
      title={Kolmogorov-Arnold Networks are Radial Basis Function Networks},
      author={Ziyao Li},
      year={2024},
      eprint={2405.06721},
      archivePrefix={arXiv},
      primaryClass={cs.LG},
      url={https://arxiv.org/abs/2405.06721},
}

@inproceedings{softs,
title={{SOFTS}: Efficient Multivariate Time Series Forecasting with Series-Core Fusion Transformer},
author={Lu Han and Xu-Yang Chen and Han-Jia Ye and De-Chuan Zhan},
booktitle={Advances in Neural Information Processing Systems},
year={2024},
url={https://openreview.net/forum?id=UMfcdRuHSt}
}

@inproceedings{timesfm,
  title={A decoder-only foundation model for time-series forecasting},
  author={Das, Abhimanyu and Kong, Weihao and Leach, Andrew and Mathur, Shaan K and Sen, Rajat and Yu, Rose},
  booktitle={Proceedings of the 41st International Conference on Machine Learning},
  year={2024},
  url={https://proceedings.mlr.press/v235/das24c.html}
}

\end{document}